\documentclass[letterpaper, 10 pt, journal, twoside]{IEEEtran}

\IEEEoverridecommandlockouts                              

\usepackage{graphicx}
\usepackage{xcolor}

\usepackage{graphics} 
\usepackage{epsfig} 
\usepackage{amsmath} 
\usepackage{amssymb}  
\usepackage{amsfonts}  
\usepackage{booktabs}  
\usepackage{siunitx}  
\usepackage[nolist,nohyperlinks]{acronym} 
\usepackage{url} 
\usepackage{mathtools} 
\usepackage{interval} 
\usepackage{bm}  
\usepackage{subcaption} 
\usepackage[implicit=false]{hyperref}
\usepackage{tabularx}
\usepackage{multirow}
\usepackage{makecell}

\usepackage{graphicx} 

\usepackage{float}

\acrodef{AAIS}[AAIS]{Autonomous Aerial Interception System}
\acrodefplural{AAIS}[AAIS']{Autonomous Aerial Interception Systems}

\acrodef{C-UAS}[C-UAS]{Counter Unmanned Aircraft System}
\acrodefplural{C-UAS}[C-UAS']{Counter Unmanned Aircraft Systems}

\acrodef{FoV}[FoV]{Field of View}
\acrodef{VFoV}[VFoV]{Vertical Field of View}
\acrodef{HFoV}[HFoV]{Horizontal Field of View}
\acrodef{RL}[RL]{Reinforcement Learning}
\acrodef{ARL}[ARL]{Application Readiness Level}
\acrodef{BFS}[BFS]{Breadth-First Search}
\acrodef{UAV}[UAV]{Unmanned Aerial Vehicle}
\acrodef{GPS}[GPS]{Global Positioning System}
\acrodef{SLAM}[SLAM]{Simultaneous Localization And Mapping}
\acrodef{SLAMs}[SLAMs]{Simultaneous Localization And Mapping systems}
\acrodef{GPS}[GPS]{Global Positioning System}
\acrodef{RTK}[RTK]{Real-time Kinematic}
\acrodef{GNSS}[GNSS]{Global Navigation Satellite System}
\acrodef{ROS}[ROS]{Robot Operating System}
\acrodef{API}[API]{Application Programming Interface}
\acrodef{UGV}[UGV]{Unmanned Ground Vehicle}
\acrodef{UV}[UV]{Ultra-Violet}
\acrodef{LED}[LED]{Light-emitting Diode}
\acrodef{MBZIRC}[MBZIRC]{Mohamed Bin Zayed International Robotics Challenge}
\acrodef{DARPA}[DARPA]{Defense Advanced Research Projects Agency}
\acrodef{SAR}[SAR]{Search and Rescue}
\acrodef{IMU}[IMU]{Inertial Measurement Unit}
\acrodef{LTI}[LTI]{Linear time-invariant}
\acrodef{MPC}[MPC]{Model Predictive Control}
\acrodef{UVDAR}[UVDAR]{Ultra-Violet Direction And Ranging}
\acrodef{DOF}[DOF]{degree-of-freedom}
\acrodef{DOFs}[DOFs]{degrees-of-freedom}
\acrodef{LiDAR}[LiDAR]{Light Detection and Ranging}
\acrodef{ESC}[ESC]{Electronic Speed Controller}
\acrodef{ICP}[ICP]{Iterative Closest Points}
\acrodef{KF}[KF]{Kalman Filter}
\acrodef{LKF}[LKF]{Linear \ac{KF}}
\acrodef{UKF}[UKF]{Unscented Kalman Filter}
\acrodef{EKF}[EKF]{Extended Kalman Filter}
\acrodef{RAS}[RAS]{Robotics and Automation Society}
\acrodef{IEEE}[IEEE]{Institute of Electrical and Electronics Engineers}
\acrodef{MRS}[MRS]{Multi-robot Systems Group}
\acrodef{CNN}[CNN]{Convolutional Neural Network}
\acrodef{CTU}[CTU]{Czech Technical University}
\acrodef{UPenn}[UPenn]{University of Pennsylvania}
\acrodef{NYU}[NYU]{New York University}
\acrodef{FIFO}[FIFO]{First In, First Out}
\acrodef{RMSE}[RMSE]{Root Mean Square Error}
\acrodef{PDF}[PDF]{Probability Distribution Function}
\acrodef{CDF}[CDF]{Cumulative Distribution Function}
\acrodef{MC}[MC]{Monte-Carlo}
\acrodef{TSDF}[TSDF]{Truncated Signed Distance Field}
\acrodef{SITL}[SITL]{Software In The Loop}

\acrodef{PP}[PP]{Pure Pursuit}
\acrodef{PN}[PN]{Proportional Navigation}
\acrodef{PPN}[PPN]{Pure Proportional Navigation}
\acrodef{TPN}[TPN]{True Proportional Navigation}
\acrodef{LPN}[LPN]{Linearized Proportional Navigation}

\acrodef{FOV}[FOV]{Field Of View}
\acrodef{LOS}[LOS]{Line Of Sight}
\acrodef{ZEM}[ZEM]{Zero-Effort Miss}
\acrodef{SWaP}[SWaP]{Size, Weight, and Power}

\acrodef{CV}[CV]{Constant Velocity}
\acrodef{CA}[CA]{Constant Acceleration}
\acrodef{CJ}[CJ]{Constant Jerk}
\acrodef{IMM}[IMM]{Interacting Multiple Model}
\acrodef{EPN}[FRPN]{Fast Response Proportional Navigation}
\acrodef{GPN}[GPN]{General Proportional Navigation}
\acrodef{VoFOD}[VoFOD]{Volumetric Flying Object Detector}

\newcommand{\norm}[1]{\left\lVert#1\right\rVert}

\usepackage[e]{esvect}

\renewcommand{\vec}[1]{\bm{#1}}
\newcommand{\pnt}[1]{\bm{#1}}
\newcommand\mat{\mathbf}

\newcommand{\bemat}[1]
{
  \begin{bmatrix}
    #1
  \end{bmatrix}
}

\newcommand{\besmat}[1]
{
  \left[
  \begin{smallmatrix}
    #1
  \end{smallmatrix}
  \right]
}

\newcommand{\tstep}[2][]{_{#1[#2]}}
\newcommand*{\tran}{^{\intercal}}

\DeclareSIUnit{\pixel}{px}
\DeclareSIUnit{\fps}{FPS}

\DeclareMathOperator*{\argmin}{argmin}

\DeclareMathOperator{\diag}{diag}

\intervalconfig{soft open fences}

\let\originalleft\left
\let\originalright\right
\renewcommand{\left}{\mathopen{}\mathclose\bgroup\originalleft}
\renewcommand{\right}{\aftergroup\egroup\originalright}

\newcommand{\set}[1]{\mathcal{\expandafter\MakeUppercase\expandafter{#1}}}

\newcommand{\at}[2]{\left.\kern-\nulldelimiterspace#1\right|_{#2}}

\usepackage{tikz}
\usepackage{scalerel}
\usetikzlibrary{svg.path}
\definecolor{orcidlogocol}{HTML}{A6CE39}
\tikzset{
  orcidlogo/.pic={
    \fill[orcidlogocol] svg{M256,128c0,70.7-57.3,128-128,128C57.3,256,0,198.7,0,128C0,57.3,57.3,0,128,0C198.7,0,256,57.3,256,128z};
    \fill[white] svg{M86.3,186.2H70.9V79.1h15.4v48.4V186.2z}
    svg{M108.9,79.1h41.6c39.6,0,57,28.3,57,53.6c0,27.5-21.5,53.6-56.8,53.6h-41.8V79.1z M124.3,172.4h24.5c34.9,0,42.9-26.5,42.9-39.7c0-21.5-13.7-39.7-43.7-39.7h-23.7V172.4z}
    svg{M88.7,56.8c0,5.5-4.5,10.1-10.1,10.1c-5.6,0-10.1-4.6-10.1-10.1c0-5.6,4.5-10.1,10.1-10.1C84.2,46.7,88.7,51.3,88.7,56.8z};
  }
}

\newcommand\orcidicon[1]{\href{https://orcid.org/#1}{\mbox{\scalerel*{
        \begin{tikzpicture}[yscale=-1,transform shape]
          \pic{orcidlogo};
        \end{tikzpicture}
}{|}}}}

\title{Towards Safe Mid-Air Drone Interception:\\Strategies for Tracking \& Capture
}

\author{Michal Pliska$^{1}$$^{\orcidicon{0009-0006-7205-0455}}$, Matouš Vrba$^{1}$$^{\orcidicon{0000-0002-4823-8291}}$, Tomáš Báča$^{1}$$^{\orcidicon{0000-0001-9649-8277}}$, and Martin Saska$^{1}$$^{\orcidicon{0000-0001-7106-3816}}$
\thanks{Manuscript received March 27, 2024; Revised July 15, 2024; Accepted August 12, 2024.}
\thanks{This paper was recommended for publication by Editor Giuseppe Loianno upon evaluation of the Associate Editor and Reviewers' comments.}
\thanks{This work was funded by CTU grant no SGS23/177/OHK3/3T/13, by the Czech Science Foundation (GAČR) under research project no. 23-07517S, and by the European Union under the project Robotics and advanced industrial production (reg. no. CZ.02.01.01/00/22\_008/0004590).}
\thanks{$^{1}$The authors are with the Faculty of Electrical Engineering, Czech Technical University in Prague, 16627 Prague 6, Czech Republic (e-mail: \texttt{\{michal.pliska, matous.vrba, tomas.baca, martin.saska\}@fel.cvut.cz}).}%
\thanks{Digital Object Identifier (DOI): see top of this page.}%
}

\newcommand{\changed}[1]{#1}

\def\x{{\bm x}}
\def\P{{\bm{\mathrm{P}}}}
\newcommand{\PREPRINTYEAR}{2024}
\newcommand{\PUBLISHEDIN}{IEEE Robotics and Automation Letters}
\newcommand{\DOI}{10.1109/LRA.2024.3451768} 

\usepackage[placement=top,vshift=-7]{background}
\SetBgScale{1.0}
\SetBgContents{\PUBLISHEDIN. PREPRINT VERSION - DO NOT DISTRIBUTE. \href{https://doi.org/\DOI}{DOI \DOI}}
\SetBgColor{black}
\SetBgAngle{0}
\SetBgOpacity{1.0}

\begin{document}


\pagenumbering{gobble}
\thispagestyle{empty}
\onecolumn
{
  \topskip0pt
  \vspace*{\fill}
  \centering
  \LARGE{%
    \copyright{} \PREPRINTYEAR~\PUBLISHEDIN\\\vspace{1cm}
    Personal use of this material is permitted.
    Permission from \PUBLISHEDIN~must be obtained for all %
other uses, in any current or future media, including reprinting or republishing this material for advertising or promotional purposes, creating new collective works, for resale or redistribution to servers or lists, or reuse of any copyrighted component of this work in other works.}
    \vspace*{\fill}
}
\NoBgThispage

\twocolumn          	
\BgThispage

\pagenumbering{arabic}
\maketitle


\begin{abstract}

  A unique approach for mid-air autonomous aerial interception of non-cooperating \aclp{UAV} by a flying robot equipped with a net is presented in this paper.
  A novel interception guidance method dubbed \acf{EPN} is proposed, designed to catch agile maneuvering targets while relying on onboard state estimation and tracking.
  The proposed method is compared with state-of-the-art approaches in simulations using \num{100} different trajectories of the target with varying complexity comprising almost \SI{14}{\hour} of flight data, and \ac{EPN} demonstrates the shortest response time and the highest number of interceptions, which are key parameters of agile interception.
  To enable a robust transfer from theory and simulation to a real-world implementation, we aim to avoid overfitting to specific assumptions about the target, and to tackle interception of a target following an unknown general trajectory.
  Furthermore, we identify several often overlooked problems related to tracking and estimation of the target's state that can have a significant influence on the overall performance of the system.
  We propose the use of a novel state estimation filter based on the \acl{IMM} filter and a new measurement model.
  Simulated experiments show that the proposed solution provides significant improvements in estimation accuracy over the commonly employed \acl{KF} approaches when considering general trajectories.
  Based on these results, we employ the proposed filtering and guidance methods to implement a complete autonomous interception system, which is thoroughly evaluated in realistic simulations and tested in real-world experiments with a maneuvering target going far beyond the performance of any state-of-the-art solution.

\end{abstract}


\vspace{-0.7em}
\section*{Supplementary Material}
{\small
\vspace{-0.3em}
\noindent \textbf{Video and data:} \url{https://mrs.felk.cvut.cz/towards-interception}
\vspace{-0.7em}}

\section{Introduction}
\begin{figure}
   \centering
   \includegraphics[width=0.48\textwidth]{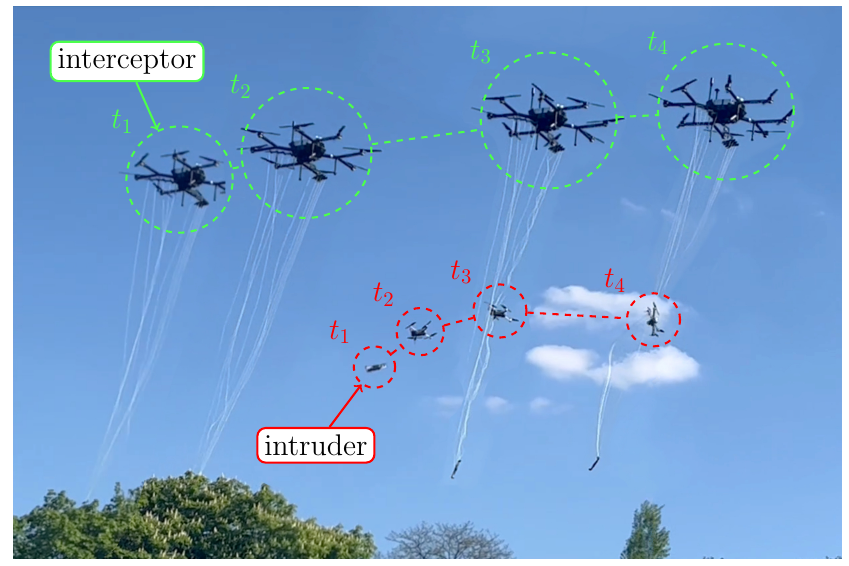}
   \caption{Collage of a successful autonomous interception of a moving target using the proposed system.
   The maneuver took approx. \SI{2}{\second} from $t_1$ to $t_4$.}
    \label{fig:intro}
\end{figure}

\IEEEPARstart{I}{nterdiction} of \acp{UAV} intruding in a restricted airspace is an important topic in aerial safety, as evidenced by many recent \ac{UAV}-related incidents and a growing research interest~\cite{chamola2021c-uas_survey, wang2021c-uas_survey, ghasri2021accidents}.
In this work, we aim to tackle autonomous aerial physical interception by a specialized robotic system with a catching net, as illustrated in Fig.~\ref{fig:intro}.
In comparison to other interdiction methods, which are mostly ground-based and/or manual, this approach offers several advantages, including the possibility of tracking and safely capturing fully autonomous and agile targets, the reduction of risk for collateral damage, and the lack of need for a skilled human operator in case of piloted systems.
However, autonomous aerial interception also presents significant technical and research challenges.
\changed{Some works have demonstrated an ability to capture a hovering target by a launched net~\cite{vrba2019ral, ning2024RealtoSimtoRealApproachVisionBased}, but a fully working system that could reliably intercept fast maneuvering targets in real-world conditions has not yet been shown in the published literature.
A net launcher only offers a limited amount of interception attempts and has also other limitations, such as the maximal flight velocity for a safe launch, which is why we choose to use a hanging net.}

The complexity of the problem is illustrated by the MBZIRC 2020 robotic competition, where one of the challenges was inspired by the aerial interception problem\footnote{\href{https://mrs.felk.cvut.cz/mbzirc2020}{\url{https://mrs.felk.cvut.cz/mbzirc2020}}}.
From 215 registered teams (22 expert international competitors selected for finals), only four teams were able to intercept the target, while relying on strong assumptions about the target's appearance and trajectory given in advance by the competition.
Inspired by our own solution~\cite{vrba2022ras}, which placed second in this challenge and first in the finals of the entire competition, we have developed a 3D \ac{LiDAR}-based detection system that removes some of these assumptions and provides accurate detection, localization, and multi-target tracking of flying objects in real-world scenarios, as described in our recent work~\cite{vrba2023fod}.
In this paper, we build upon that onboard perception and propose a novel interception guidance technique.
This, together with the proposed advanced state estimation and trajectory prediction of the target relying on minimal assumptions about the target's trajectory, enables us to reach a full system capable of real-world deployment.

\vspace{-2pt}
\subsection{Related Works}

The two main research domains relevant to designing an \ac{AAIS} are detection and tracking of the target, and guidance of the interceptor towards the target.
The detection and tracking of intruder \acp{UAV} typically include also estimation of the target's state that serves as input for the subsequent guidance algorithm, and thus forms the foundation of an effective interception.
Assuming a well-estimated state of the target, the system's capability to intercept it is then determined by the navigation algorithm.
An overview of related works in these two fields is provided in this section.


\subsubsection{Detection and Tracking}
\label{sec:rw_dettrack}
Most of the research on onboard detection of non-cooperating \acp{UAV} focuses on visual methods based on deep learning \cite{barisic2022fr, vrba2020ral, nguyen2019mavnet}.
However, detection and precise state estimation of small drones in cluttered environments proves to be a difficult problem, which several works attempt to address by utilizing depth images instead of (or in addition to) RGB~\cite{vrba2019ral, carrio2018depth}.
Flying objects are typically spatially distinctive, making them easily detectable in depth images, and the depth information provides a full 3D relative location of the target, which is otherwise problematic with monocular approaches.
Our recent research~\cite{vrba2023fod} further builds upon this idea and takes advantage of the high accuracy of 3D \ac{LiDAR} sensors to build a volumetric occupancy map of the environment and detect flying objects within this map, which provides more context and thus enables detection with a minimum of false positives even in complex environments.

Accurate state estimation and trajectory prediction of the target is crucial for agile interception.
In~\cite{ning2024bearing}, an approach for bearing-only position and velocity estimation of a visually-tracked target based on a modified \ac{KF} is proposed.
If the full 3D position of the target is measured, a \ac{CV} or \ac{CA} \ac{KF} is typically employed, such as in~\cite{vrba2023fod, barisic2022fr,srivastava2022intercption}.
However, we have observed that assuming a \ac{CV} or even a \ac{CA} dynamic model proves insufficient to predict trajectories of targets following non-trivial trajectories.
In this work, we rely on the output of the detection and tracking system from~\cite{vrba2023fod}, and go beyond that solution by integrating an \ac{IMM} filter, which offers better accuracy in the considered cases, as confirmed by our experiments.



\subsubsection{Navigation for UAV Interception}
The simplest approach to interception guidance is \ac{PP}, which only considers the target's current relative position.
This approach is used in~\cite{wang2020multi_interceptor}, where a cooperative multi-UAV interception strategy is proposed.
A more sophisticated method is \ac{PN}, which is a commonly used technique originally developed for missile guidance that generally provides more accurate interception than \ac{PP}.
In~\cite{moreira2019interception}, \ac{PN}-based guidance combined with a \ac{CA} model of the target is designed and evaluated on three trajectories in real-world experiments with a virtual \ac{UAV} target.
A major limitation of \ac{PN} applied to \acp{UAV}, is identified in~\cite{zhu2017DistributedGuidanceInterception}.
For the \ac{PN} guidance law to converge towards interception, the relative velocity between the interceptor and the target must be negative.
The authors propose a modification dubbed \ac{GPN} to address this limitation.
The interception planning problem can also be formulated using \ac{MPC} optimization as in~\cite{srivastava2022intercption}, where the interceptor's desired acceleration and heading is selected by minimizing the Euclidean distance between the predicted trajectory of the target and the interceptor.
However, the authors of~\cite{srivastava2022intercption} assume a car-like model of the interceptor \ac{UAV}, which is overly restrictive and leads to a reduced performance.
\changed{%
An often overlooked problem is a limited field of view of real sensors.
This is tackled in~\cite{yang2023PolicyLearning}, where a \ac{RL}-based planning strategy for tracking multiple targets using agents with limited field of view is presented. 
Although \ac{RL} is a promising field in motion planning, it is yet to be applied to aerial interception.}

\changed{Interception using a launched net is a similar task to the one in this paper, but it has different navigation requirements.
This is tackled in~\cite{vrba2019ral} using \ac{PP}, and in~\cite{ning2024RealtoSimtoRealApproachVisionBased} using a custom guidance law that also takes into account velocity of the target.
Both works demonstrate a real-world capture of a hovering target using these techniques while relying on feedback from onboard cameras, putting them the closest to a fully working \ac{AAIS}.
However, such navigation is not easily transferable to agile interception using a hanging net.}

\vspace{-0.5em}

\subsection{Contributions}

\changed{
In the context of the state of the art, we summarize our contributions as follows:
\begin{enumerate}
  \item Two new methods for guiding an \ac{AAIS} \ac{UAV} to physically intercept another \ac{UAV} are proposed: an \ac{MPC}-based trajectory planner, and a \ac{PN}-based \acf{EPN} guidance law.
  \item A detailed comparison of several approaches to state estimation of the target is provided, including a novel measurement uncertainty model.
  \item The methods are integrated into a full \ac{AAIS} solution and compared with state-of-the-art state estimation and guidance algorithms.
    The proposed methods offer a significant improvement in time to first contact with the target and in the number of successful interceptions.
  \item Our dataset of testing trajectories is publicly available to enable reproducing our results and to enable comparison with future new interception navigation methods.
  \item Finally, viability of the designed \ac{AAIS} system is demonstrated in several real-world experiments, showing the capability to intercept a target flying \SI{5}{\metre\per\second}.
\end{enumerate}
}

\vspace{-0.5em}

\section{State Estimation and Prediction}

As mentioned in the previous sections, we rely on the \ac{LiDAR}-based \ac{VoFOD} algorithm presented in~\cite{vrba2023fod} to detect the target \ac{UAV}, and on the multi-target tracking algorithm from the same work.
To estimate the state of the target and predict its motion, \ac{VoFOD} relies on a standard \ac{KF} and assumes a \ac{CA} motion model.
Although most typical \ac{UAV} movement patterns consist of waypoint-following or similar trivial trajectories for which a well-tuned \ac{CV} or \ac{CA} model is sufficient, in this work, we aim to design a solution that works more universally also for agile maneuvering targets.
As the results in sec.~\ref{sec:eval_tracking} show, a suitably formulated and well-tuned \ac{IMM} filter provides superior performance in this regard.


\subsection{Interacting Multiple Models Filter}\label{imm}
\label{sec:imm}



An \ac{IMM} filter consists of \(m\) estimators running in parallel, each assuming a different dynamic model. Their estimates are combined based on the assumption that the transitions between the dynamic models follow a Markov chain model.
This is done in two phases: a \textit{filtering} phase, and a \textit{mixing} phase.

\changed{%
The filter estimates the likelihood $\mu^j\tstep{k}$ of each model.
Here, $\tstep{k}$ denotes the time step and $j$ the $j$-th model.
In the \textit{filtering} phase, the separate estimators are updated using their respective models and a measurement to obtain state estimates $\hat{\x}\tstep{k-}^{j}$, and the corresponding covariances $\P\tstep{k-}^j$.
In the \textit{mixing} phase, the models interact and are re-initialized and re-weighted.
For the $j$-th filter, this is performed as
\begin{align}
  \hat{\x}\tstep{k}^{j} &= \sum_{i=1}^{m} \mu\tstep{k}^{i \mid j}\hat{\x}\tstep{k-}^i, \label{eq:mixing_state} \\
  \P\tstep{k}^{j} &= \sum_{i=1}^{m} \mu\tstep{k}^{i \mid j} \Big( \P\tstep{k-}^i + \left[ \hat{\x}\tstep{k}^i - \hat{\x}\tstep{k}^{j} \right] \left[ \hat{\x}\tstep{k}^i - \hat{\x}\tstep{k}^{j} \right]\tran \Big), \label{eq:mixing_cov}
\end{align}
where $\hat{\x}\tstep{k}^{j}$ is the state estimate after mixing, $\P\tstep{k}^{j}$ is the corresponding covariance, and \(\mu\tstep{k}^{i \mid j}\) is the conditional probability of model \(i\) transitioning to model \(j\).
The transition probability $\mu\tstep{k}^{i \mid j}$ is computed using the measurement innovations of the individual filters and parameters of the Markov chain model.
The overall estimate \(\hat{\x}\tstep{k}\) and covariance \(\P\tstep{k}\) are then obtained using the same eqs.~\eqref{eq:mixing_state},~\eqref{eq:mixing_cov} with \(\mu\tstep{k}^{i}\) substituted for \(\mu\tstep{k}^{i \mid j}\).
For a more thorough description of \ac{IMM}, we refer the reader to~\cite{rong2005imm_survey}.}


In this work, we propose using an \ac{IMM} filter with two motion models, \ac{CV} and \ac{CA}, and a position measurement of the target, which is provided by \ac{VoFOD}.
The dynamic models and the measurement model are formulated as stochastic discrete state-space systems defined as
\begin{align}
  \pnt{x}\tstep{k+1} &= \mat{A}\tstep{k}\pnt{x}\tstep{k} + \pnt{\xi}\tstep{k}, && \pnt{\xi}\tstep{k} \sim \mathcal{N}\left(\pnt{0}, \mat{\Xi}\tstep{k}\right), \label{eq:ss} \\
  \pnt{z}\tstep{k} &= \mat{H}\pnt{x}\tstep{k} + \pnt{\zeta}\tstep{k}, && \pnt{\zeta}\tstep{k} \sim \mathcal{N}\left(\pnt{0}, \mat{Z}\tstep{k}\right), \label{eq:meas}
\end{align}
where $\pnt{x}$ is a state vector, $\mat{A}$ is a state transition matrix, $\pnt{\xi}$ is a Gaussian process noise with zero mean and a covariance matrix $\mat{\Xi}$, $\pnt{z}$ is a measurement, $\mat{H}$ is a measurement matrix, and $\pnt{\zeta}$ is a Gaussian measurement noise with covariance $\mat{Z}$.
The lower index $\tstep{k}$ will further be omitted unless relevant for brevity.
The specific \ac{CV} and \ac{CA} models are described below, and are updated in the \textit{filtering} phase as standard \acp{KF}.


\subsubsection{Constant Velocity Dynamic Model}
\label{sec:cv}

For the \ac{CV} motion model, the variables from eqs.~\eqref{eq:ss} and~\eqref{eq:meas} are defined as
\begin{align}
  \pnt{x} &= \bemat{ \pnt{p}\tran & \pnt{v}\tran }\tran, &&
  \mat{A} = \besmat{
    \mat{I} & \Delta t \mat{I} \\
    \mat{0} & \mat{I}
  }, \\
  \pnt{\xi} &= \mat{B} \pnt{a}, &&
  \mat{B} = \besmat{
    \frac{1}{2} \Delta t^2 \mat{I} \\
    \Delta t \mat{I}
  }, \\
  \pnt{a} &\sim \mathcal{N}\left(\pnt{0}, \sigma_{\pnt{a}}^2 \mat{I}\right), &&
  \mat{H} = \bemat{
    \mat{I} & \mat{0}
  },
\end{align}
where $\pnt{p} \in \mathbb{R}^3$ and $\pnt{v} \in \mathbb{R}^3$ denote the target's position and velocity, respectively, $\mat{I} \in \mathbb{R}^{3\times 3}$ is an identity matrix, $\mat{0} \in \mathbb{R}^{3\times 3}$ is a zero matrix, $\Delta t \in \mathbb{R}$ is the time-step duration, $\mat{B} \in \mathbb{R}^{6\times 3}$ is an input matrix, and $\pnt{a} \in \mathbb{R}^3$ is an unknown acceleration of the target that is assumed to be a Gaussian random variable with zero mean and covariance $\sigma_{\pnt{a}}^2\mat{I}$.
The process noise covariance matrix is thus $\mat{\Xi} = \sigma_{\pnt{a}}^2 \mat{B}\mat{B}\tran$.
The measurement covariance matrix $\mat{Z}$ is obtained using the method described in sec.~\ref{sec:meas}.



\subsubsection{Constant Acceleration Dynamic Model}
\label{sec:ca}

The \ac{CA} motion model is described similarly to the \ac{CV} model as
\begin{align}
  \pnt{x} &= \bemat{ \pnt{p}\tran & \pnt{v}\tran & \pnt{a}\tran }\tran, &&
  \mat{A} = \besmat{
    \mat{I} & \Delta t \mat{I} & \frac{1}{2}\Delta t^2\mat{I} \\
    \mat{0} & \mat{I} & \Delta t \mat{I} \\
    \mat{0} & \mat{0} & \mat{I} \\
  }, \label{eq:CA_state} \\
  \pnt{\xi} &= \mat{B} \pnt{j}, &&
  \mat{B} = \besmat{
    \frac{1}{6} \Delta t^3 \mat{I} \\
    \frac{1}{2} \Delta t^2 \mat{I} \\
    \Delta t \mat{I}
  }, \label{eq:CA_input} \\
  \pnt{j} &\sim \mathcal{N}\left(\pnt{0}, \sigma_{\pnt{j}}^2 \mat{I}\right), &&
  \mat{H} = \bemat{
    \mat{I} & \mat{0} & \mat{0}
  },
\end{align}
where $\pnt{j} \in \mathbb{R}^3$ is an unknown jerk of the target that is assumed to be a Gaussian random variable with zero mean and covariance $\sigma_{\pnt{j}}^2\mat{I}$.
The process covariance is thus $\mat{\Xi} = \sigma_{\pnt{j}}^2 \mat{B}\mat{B}\tran$.




\subsection{LiDAR Measurement Uncertainty}
\label{sec:meas}

For an accurate estimation of the target's state, an appropriate model of the measurement uncertainty is crucial.
In the used motion models, the measurement $\pnt{z} \in \mathbb{R}^3$ is defined as
\begin{align}
  \vec{z} = \vec{p} + \vec{\zeta},
\end{align}
where $\pnt{p}$ is the target's actual position in a static world frame $\mathcal{W}$, and $\vec{\zeta}$ is noise.
Following the results from~\cite{vrba2023fod}, we assume that there are three major sources of uncertainty in $\vec{z}$:
\begin{enumerate}
  \item the range of the target $l_m$ measured by the \ac{LiDAR},
  \item the observer's own pose used to transform the \ac{LiDAR} points from the observer's frame $\mathcal{O}$ to $\mathcal{W}$, which we define as a transformation $\mat{T}_{\mathcal{O}}^{\mathcal{W}}$,
  \item biased sampling of the measured points of the target, which is observed from one direction.
\end{enumerate}

For the first two uncertainties, we adopt the model from~\cite{vrba2023fod} and approximate the result as an additive Gaussian noise $\vec{\zeta}_{12} \in \mathbb{R}^3$, defined as
\begin{align}
  \vec{\zeta}_{12} \sim \mathcal{N}\left( \vec{0}, \mat{\Sigma}_{\vec{\zeta}_{12}} \right),
\end{align}
where the covariance $\mat{\Sigma}_{\vec{\zeta}_{12}} \in \mathbb{R}^{3\times 3}$ is a function of the pose $\mat{T}_{\mathcal{O}}^{\mathcal{W}}$, the target's measured range $l_m$, and their uncertainties expressed as a covariance $\mat{\Sigma}_{\mat{T}}$ and standard deviation $\sigma_l$.
Based on the results from~\cite{vrba2023fod}, we propose to model the third uncertainty as an additive Gaussian noise $\vec{\zeta}_3 \in \mathbb{R}^3$, defined as
\begin{align}
  \vec{\zeta}_3 \sim \mathcal{N}\left( \vec{0}, \mat{\Sigma}_{\vec{\zeta}_3} \right),
\end{align}
where $\mat{\Sigma}_{\vec{\zeta}_3}\in \mathbb{R}^{3\times 3}$ is the covariance matrix.
The probability distribution of the total measurement noise $\vec{\zeta} = \vec{\zeta}_{12} + \vec{\zeta}_3$ of the target's position $\vec{z}$ is then
\begin{align}
  \vec{\zeta} \sim \mathcal{N}\left( \vec{0}, \mat{\Sigma} \right) = \mathcal{N}\left( \vec{0}, \mat{\Sigma}_{\vec{\zeta}_{12}} + \mat{\Sigma}_{\vec{\zeta}_3} \right).
\end{align}

In practice, the uncertainty $\sigma_l$ of the range measurement $l_m$ is provided by the sensor manufacturer, and the error caused by the sampling bias is typically chosen as $\mat{\Sigma}_{\vec{\zeta}_3} = \sigma_{\vec{\zeta}_3} \mat{I}$ with $\sigma_{\vec{\zeta}_3}$ as an empirically determined parameter.
To select the observer's pose uncertainty covariance $\mat{\Sigma}_{\mat{T}}$, we assume that the translation and rotation errors are independent, and that $\mat{\Sigma}_{\mat{T}}$ can thus be expressed as
\begin{align}
  \mat{\Sigma}_{\mat{T}} = \besmat{
    \mat{\Sigma}_{\vec{t}} & \mat{0} \\
    \mat{0} & \mat{\Sigma}_{\vec{\alpha}}
  },
\end{align}
where $\mat{\Sigma}_{\vec{t}} \in \mathbb{R}^{3\times 3}$ is the covariance of the translation error $\vec{t}_e \in \mathbb{R}^3$, and $\mat{\Sigma}_{\vec{\alpha}} \in \mathbb{R}^{3\times 3}$ is the covariance of the rotation error $\vec{\alpha}_e \in \mathbb{R}^3$, with $\vec{\alpha}_e = \bemat{\alpha_e, \beta_e, \gamma_e}\tran$ representing rotations around the $X$, $Y$, and $Z$ axes, respectively.
\changed{%
It is important to note that the rotation uncertainty is expressed in the body frame of the observer, so the angles are assumed to be close to zero.
The distribution is therefore linearized around $\vec{\alpha}_e = \vec{0}$, so there is no singularity of the representation.
We choose this model because it is intuitive to interpret and corresponds well to the observed behavior of the system.}

The covariance $\mat{\Sigma}_{\vec{t}}$ is reported by most self-localization systems, including the \ac{RTK} \ac{GPS} module used during our experiments.
However, obtaining $\mat{\Sigma}_{\vec{\alpha}}$ is problematic, yet it can have a significant influence on the resulting total uncertainty.
We observe, that for the widely adopted PixHawk 4 flight controller used in our system for attitude control and orientation estimation, $\mat{\Sigma}_{\vec{\alpha}}$ can be estimated as a function of angular velocities $\vec{\omega} \in \mathbb{R}^3$ around the corresponding axes as
\begin{align}
  \mat{\Sigma}_{\vec{\alpha}} = c_{\vec{\alpha}}^2 \diag\left( \vec{\omega}^2 \right),
\end{align}
where $c_{\vec{\alpha}}$ is an empirically determined parameter \changed{and $\vec{\omega}^2$ denotes an element-wise exponentiation of $\vec{\omega}$}.




\section{Interception Navigation and Planning}
In this section, we present the methods used for navigation of the interceptor towards interception of the target.
Firstly, we discuss variants of \acf{PN} that motivated the design of the proposed \ac{EPN} approach introduced in the second part.
The last part presents a novel interception trajectory planner formulated using \acf{MPC} reflecting the requirements of agile interception.

\subsection{Proportional Navigation}
\label{sec:pn_method}

\begin{figure}
  \centering
  \includegraphics[width=1\columnwidth]{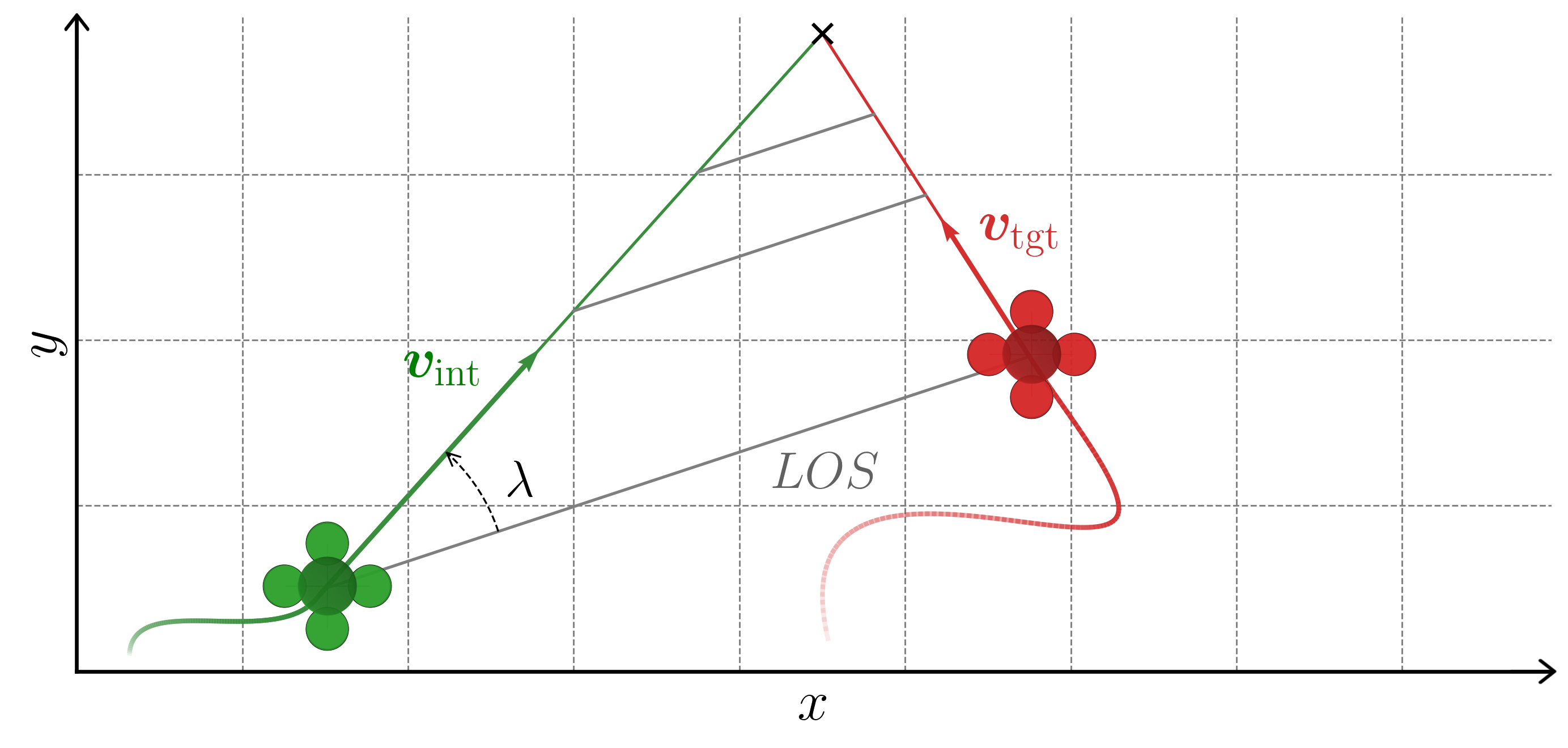}
  \caption{Illustration of the fundamental concepts of \acf{PN}, highlighting the \acf{LOS}. \(\vec{v}_{\text{int}}\) denotes the velocity of the interceptor, and \(\vec{v}_{\text{tg}}\) denotes the velocity of the target.}
  \label{fig:pn_concept}
\end{figure}

\acf{PN} is a guidance technique widely used in homing missile technology~\cite{missels}.
The central principle of \ac{PN} involves the \acf{LOS}, defined as a line connecting the interceptor and the target (see Fig.~\ref{fig:pn_concept}).
If the interceptor and the target have a constant velocity and are on a collision course, the \ac{LOS} and the lines connecting the interceptor and the target with the collision point form a so-called collision triangle.
In such a case, it may be observed that the angle between the \ac{LOS} and the interceptor's velocity is constant.
This angle is commonly called the \ac{LOS} angle and denoted $\lambda$.
\ac{PN} aims to reach this state by controlling the interceptor's acceleration to maintain a constant \ac{LOS} angle.


The canonical \ac{PN} control rule can be formulated as
\begin{equation}
  \vec{a}_{\text{cmd}} = G \cdot v_{\text{c}} \cdot \dot{\lambda} \cdot \vec{a}_{\text{dir}}, \label{eq:pn_gen}
\end{equation}
where $\vec{a}_{\text{cmd}} \in \mathbb{R}^3$ is the commanded acceleration, $G \in \mathbb{R}$ is control gain, $v_{\text{c}} \in \mathbb{R}$ is the closing velocity, $\dot{\lambda} \in \mathbb{R}$ is the rate of change of the \ac{LOS} angle, and $\vec{a}_{\text{dir}} \in \mathbb{R}^3$ is a unit vector specifying the acceleration direction.
The concrete definition of $v_{\text{c}}$ and $\vec{a}_{\text{dir}}$ depends on the specific variation of \ac{PN} used.

However as acknowledged in~\cite{zhu2017DistributedGuidanceInterception}, this formulation has a major drawback in the tackled application.
The \ac{LOS} angle rate $\dot{\lambda}$ is zero not only when the target and the interceptor are flying along the sides of the collision triangle towards the collision point, but also when flying away from it along the same lines.
Therefore, if the interceptor misses the target and continues flying, $\dot{\lambda}$ is close to zero, resulting in a very small $\vec{a}_{\text{cmd}}$, and thus a slow response of the interceptor.
This issue is mostly irrelevant for missiles, and is therefore scarcely addressed in the literature, but it cannot be ignored for \ac{UAV} interception, where the ability to reattempt the interception is crucial.


The problem can be circumvented by adopting an alternative \ac{PN} formulation.
\changed{Assuming a constant velocity of the target, unlimited action inputs of the interceptor, and several other conditions, \ac{PN} can be derived using optimal control theory regarding the miss distance as the objective, yielding what is termed the Cartesian form or the \ac{LPN} (for more details, we refer the reader to \cite{palumbo_missiles}).}
The \ac{LPN} control rule can be formulated as
\begin{equation}
  \vec{a}_{\text{cmd}} = G \frac{\Delta \vec{p} + \Delta \vec{v} \cdot t_{\text{go}}}{t_{\text{go}}^2} , \label{eq:pn}
\end{equation}
where $\Delta \vec{p} \in \mathbb{R}^3$ is the relative position of the target, $\Delta \vec{v} \in \mathbb{R}^3$ is the relative velocity, and $t_{\text{go}} \in \mathbb{R}$ is the time to collision estimated as $t_{\text{go}} = \frac{\norm{ \Delta \vec{p} }}{ \norm{\Delta \vec{v}} }$.

\subsection{\acl{EPN} Formulation}
\begin{figure}
  \centering
  \begin{subfigure}[t]{0.48\linewidth}
    \centering
    \includegraphics[width=.777\linewidth]{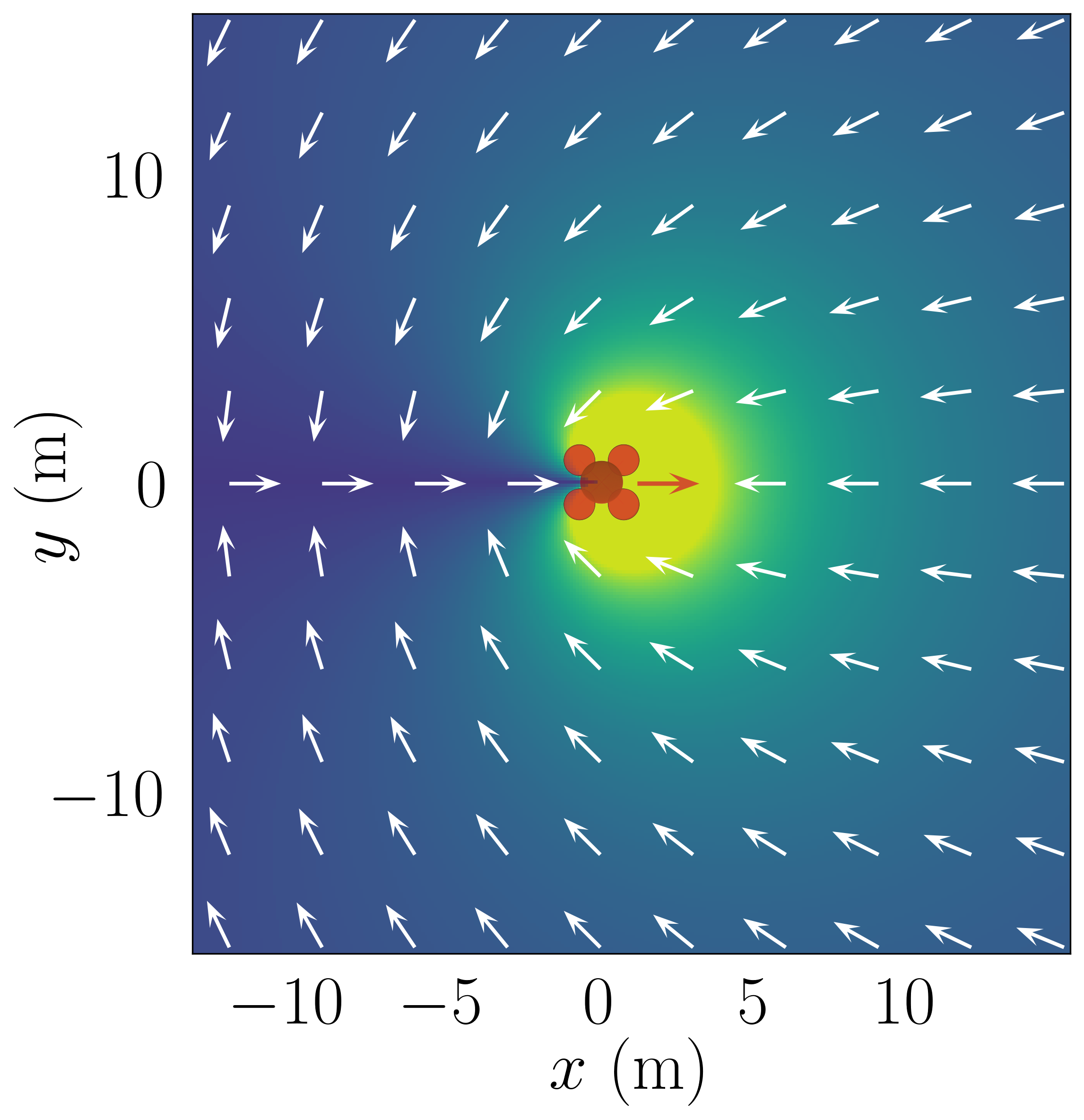}
    \caption{\acl{LPN}.}
    \label{fig:PN_ZEM}
  \end{subfigure}
  \begin{subfigure}[t]{0.48\linewidth}
    \centering
    \includegraphics[width=.868\linewidth]{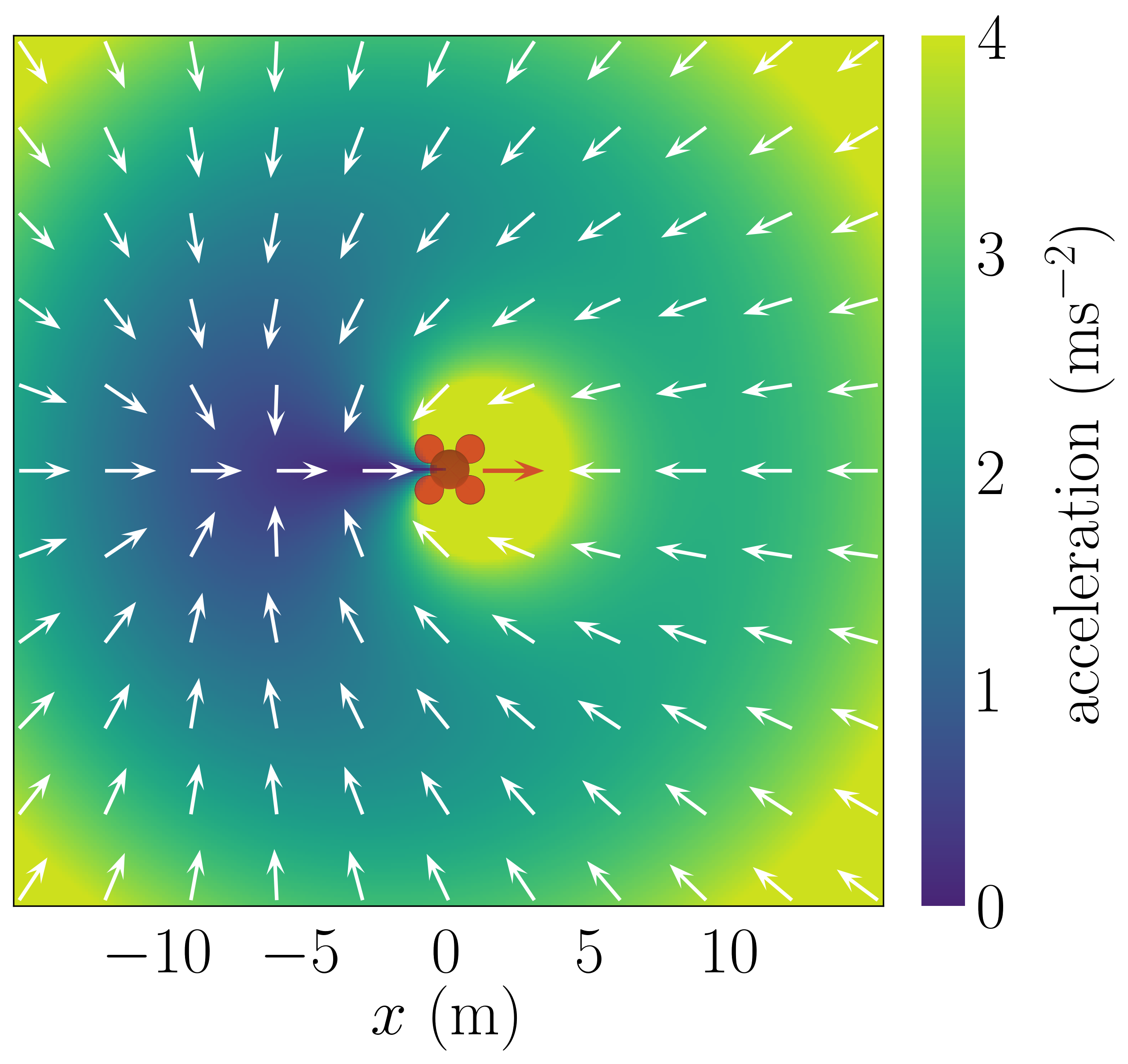}
    \caption{\acl{EPN}.}
    \label{fig:EPN_strategy}
  \end{subfigure}
  \caption{Comparison between the commanded acceleration of \ac{LPN} and \ac{EPN}.
  The white arrows and indicate the acceleration and the color its magnitude for different positions of the interceptor.
  The target is positioned in the center and is flying at \SI{1}{\metre\per\second} along the $x$-axis (red arrow).
  The interceptor is flying at \SI{2}{\metre\per\second} in the same direction.
  }
  \label{fig:LPN_EPN}
\end{figure}

The above mentioned solution avoids the divergence problem, but fails to initiate engagement when $\Delta \vec{v}$ is close to zero.
Moreover, as shown in sec.~\ref{sec:res_pn}, the engagement is slow and results in a long time to first contact.
On the other hand, the simple \ac{PP} control law produces an output proportional to the distance to the target resulting in a fast approach, but is prone to overshooting and large miss distances, which is why it is typically not employed.
We propose a new control law, termed \acf{EPN}, to provide a solution suited for the agile interception of highly maneuvering targets.
The proposed \ac{EPN} control rule is formulated as
\begin{equation}
  \vec{a}_{\text{cmd}} = G \left( (1-W) \frac{ \Delta \vec{p} + \Delta \vec{v} \cdot t_{\text{go}}}{t_{\text{go}}^2} + W \cdot \Delta \vec{p} \right) , \label{eq:epn}
\end{equation}
where $W \in \mathbb{R}$ is a weight of the newly introduced proportional approach term $\Delta\vec{p}$.
Fig.~\ref{fig:LPN_EPN} compares the commanded accelerations of \ac{LPN} and \ac{EPN} to illustrate the improved initial engagement of \ac{EPN}.
As we demonstrate in sec.~\ref{sec:res_pn}, this modification significantly improves the speed of the initial and repeated engagements, which enables more attempts of the interceptor to catch the target, at a negligible cost in accuracy.


\subsection{\acl{MPC}}

Although \acl{PN} can achieve optimality under specific conditions with respect to the miss distance, it cannot incorporate dynamic constraints of the \acp{UAV}.
To overcome this limitation, we adopt the \ac{MPC} framework and design a trajectory planning algorithm which implements a motion model of the target and the interceptor.
We model both \acp{UAV} as point masses up to the \changed{second} derivation similarly as the \ac{CA} motion model presented in sec.~\ref{sec:ca}.
Specifically, we denote the states of the interceptor and the target as
\begin{align}
  \vec{x}_{\text{int}} &= \bemat{ \vec{p}_{\text{int}}\tran & \vec{v}_{\text{int}}\tran & \vec{a}_{\text{int}}\tran }\tran, \\
  \vec{x}_{\text{tgt}} &= \bemat{ \vec{p}_{\text{tgt}}\tran & \vec{v}_{\text{tgt}}\tran & \vec{a}_{\text{tgt}}\tran }\tran.
\end{align}
The interceptor's state $\vec{x}_{\text{int}} \in \mathbb{R}^9$ is produced by its self-localization and estimation pipeline using its onboard sensors, and the target's state $\vec{x}_{\text{tgt}} \in \mathbb{R}^9$ is estimated by the filtering algorithm, as discussed in sec.~\ref{sec:imm}.
This formulation allows for prediction of the interceptor's future state $\hat{\vec{x}}_{\text{int}}$ and the target's future state $\hat{\vec{x}}_{\text{tgt}}$ using the motion model defined in the eqs.~\eqref{eq:ss}, \eqref{eq:CA_state}, and \eqref{eq:CA_input}, by setting $\vec{j} = \vec{0}$.

The objective of the \ac{MPC} is to minimize the predicted position error $\hat{\vec{e}} = \hat{\vec{p}}_{\text{tgt}} - \hat{\vec{p}}_{\text{int}}$ between the interceptor and the target over a control horizon by controlling the input $\vec{u}$, which we define as $\vec{u} = \vec{a}_{\text{int}}$.
The control horizon is discretized to $N$ steps each with duration $\Delta t$, which we denote as $k \in \left\{ 0, 1, \dots, N-1 \right\}$, with $k=0$ corresponding to the current time $t_0$, $k=1$ to $t_0 + \Delta t$, etc.
The optimal control problem is then formulated as
\begin{align}
  \mathcal{U}^* &= \argmin_{ \mathcal{U} } J\left( \mathcal{U} \right), \\
  J\left( \mathcal{U} \right) &= \sum_{k=0}^{N-1} \hat{\vec{e}}\tstep{k}\tran \mat{W}_{\vec{e}} \hat{\vec{e}}\tstep{k} + \sum_{k=0}^{N-1} \vec{u}\tstep{k}\tran \mat{W}_{\vec{u}} \vec{u}\tstep{k},
\end{align}%
\begin{alignat}{2}
  &\text{subject to}\nonumber \\
  \hat{\vec{x}}\tstep[\text{int}]{0} &= \vec{x}\tstep[\text{int}]{0}, &
  \hat{\vec{x}}\tstep[\text{tgt}]{0} &= \vec{x}\tstep[\text{tgt}]{0}, \\
  \hat{\vec{x}}\tstep[\text{int}]{k+1} &= \mat{A} \hat{\vec{x}}\tstep[\text{int}]{k}, &
  \hat{\vec{x}}\tstep[\text{tgt}]{k+1} &= \mat{A} \hat{\vec{x}}\tstep[\text{tgt}]{k}, \\
  -\vec{v}_{\text{max}} & \leq \vec{v}\tstep[\text{int}]{k} \leq \vec{v}_{\text{max}}, &
  -\vec{a}_{\text{max}} & \leq \vec{a}\tstep[\text{int}]{k} \leq \vec{a}_{\text{max}},
\end{alignat}
where $\mathcal{U} = \left\{ \vec{u}\tstep{k} \right\}$ is an input sequence of $N$ control inputs, $\mathcal{U}^*$ is the optimal input sequence, $J\left( \mathcal{U} \right)$ is the cost function, $\mat{W}_{\vec{e}} \in \mathbb{R}^{3\times 3}$ is an error weighting matrix, $\mat{W}_{\vec{u}} \in \mathbb{R}^{3\times 3}$ is an input weighting matrix, and $\vec{v}_{\text{max}}$ and $\vec{a}_{\text{max}}$ are the interceptor's dynamic constraints.
For the experimental evaluation, the optimization problem is implemented and solved using the ACADOS \ac{MPC} framework~\cite{verschueren2020acados}.


\section{Experiments}

In this section, we validate our system in simulations by addressing the following questions:
\begin{itemize}
    \item \textbf{Tracking a maneuvering target:} 
      Is the \ac{IMM} filter more effective than traditional approaches?
      What is the impact of agile maneuvering on the quality of estimation when utilizing the proposed measurement model?
    
    \item \textbf{Fast Response Proportional Navigation:} 
      How does the proposed \ac{EPN} compare with established variants of \ac{PN} in terms of effectiveness?
      How does our adjustment affect the time to first contact and interception accuracy?

    \item \textbf{Evaluation in realistic simulations:} 
      How do the considered methods perform in realistic simulations that include a fully simulated \ac{SITL} \ac{UAV} control pipeline, \ac{LiDAR}-based detection, tracking, and target's state estimation?
\end{itemize}
Finally, we demonstrate our system's practicality in real-world experiments including an autonomous capture of a maneuvering target.
The proposed methods are implemented using the MRS UAV system \cite{mrs_stack}, which provides self-localization and state estimation, trajectory tracking, and low-level control for \acp{UAV} (among other sub-systems).

\changed{%
Unless specified otherwise, the experiments were simulated in the Gazebo simulator
with the MRS X500 \ac{UAV} (see~\cite{HertJINTHW_paper}) for the interceptor and the target, which is a quadcopter based on the Holybro X500 frame with an approx. \SI{0.5}{\metre} diameter, \num{3.7} thrust-to-weight ratio, and weights \SI{3}{\kilo\gram} (incl. the battery).
The real-world experiments were realized using the custom Eagle.One\footnote{\href{https://eagle.one}{\url{https://eagle.one}}} interceptor, which is approx. \SI{1.6}{\metre} in diameter, has a \num{3.1} thrust-to-weight ratio, weights \SI{16}{\kilo\gram}, and is equipped with the Ouster OS1-128 \ac{LiDAR}, and a $\SI{2}{\metre} \times \SI{3}{\metre}$ deployable net (depicted in Fig.~\ref{fig:intro}).
Both \ac{UAV} platforms carry an \ac{RTK}-\ac{GPS} for self-localization, an Intel NUC computer running the state estimation, control, detection, and navigation, and a PixHawk 4 flight controller for low-level stabilization and control.
Videos from the experiments and the testing trajectories are available online\footnote{\href{https://mrs.felk.cvut.cz/towards-interception}{\url{https://mrs.felk.cvut.cz/towards-interception}}}.
}

\subsection{Tracking a Maneuvering Target}\label{sec:eval_tracking}

We compare the state estimation algorithms and evaluate the proposed \ac{LiDAR} measurement uncertainty model on detections of the target using the \ac{VoFOD} algorithm in a high-fidelity \ac{SITL} Gazebo simulation environment that provides a realistic simulation of the \ac{LiDAR} sensor.
\changed{%
Six trajectories of various complexity are selected from the 100 testing trajectories designed for the quantitative evaluation of navigation methods (see the next section).
Only a subset is selected for this evaluation because simulating the full set manually is infeasible (Gazebo does not provide an easy way to automate such task).}
For each trajectory tracked by the target, three sets of data are captured with the interceptor first \textit{hovering}, then slowly \textit{flying} (with body rotation rates $\leq \SI{1}{\radian\per\second}$), and then \textit{maneuvering} (rotation rates up to $\SI{2.3}{\radian\per\second}$).
The target achieves velocities up to \SI{8}{\metre\per\second} and accelerations up to \SI{4}{\metre\per\second\squared} during these experiments.
Each experiment lasts approx. \SI{90}{\second}, and the detections have a rate of \SI{10}{\hertz}, which results in almost \SI{30}{\minute} of data and around \num{17000} detection samples in total.

The performance of the state estimators is significantly influenced by the selection of their parameters, and manual calibration of these methods is susceptible to bias and oversights.
Although there exist methods for tuning \acp{KF}, systematic tuning of the \ac{IMM} filter is not well-established.
To provide a fair comparison of the different algorithms, we tune the filters through numerical optimization by minimizing the \ac{RMSE} over the measured data using the Broyden–Fletcher–Goldfarb–Shanno algorithm.

\changed{%
Specifically, the optimization metric is defined as
\begin{equation}
  e_{\vec{x}} = \sqrt{\frac{1}{n} \sum_{i=1}^{n} \norm{ \vec{p}_i - \vec{p}_i^{\text{gt}} }_2^2 + c_e^2 \norm{ \vec{v}_i - \vec{v}_i^{\text{gt}} }_2^2},
\end{equation}
where $n$ is the total number of samples, $\vec{p}_i$, and $\vec{v}_i$ are the $i$-th estimated position and velocity of the target, $\vec{p}_i^{\text{gt}}$, and $\vec{v}_i^{\text{gt}}$ are the corresponding ground truth, and $c_e = \SI{1}{\second}$ is a unit normalization constant.}
This metric is also used to compare the performance of the estimators.
We also separately compare the position error $e_{\vec{p}}$ and the velocity error $e_{\vec{v}}$ (defined analogously to $e_{\vec{x}}$).
The results are presented in Table~\ref{tab:tracking}.


{
\sisetup{round-mode=places,round-precision=3,per-mode=repeated-symbol}
\newcolumntype{L}{>{\raggedright\arraybackslash}X}%
\begin{table}
  \vspace{0.8em}
\centering
  \begin{tabularx}{\linewidth}{lLLLLLL}
\toprule
  \multirow{2}{5em}{\textbf{Interceptor mode}} & \multicolumn{3}{c}{\textbf{No measurement model}} & \multicolumn{3}{c}{\textbf{With measurement model}} \\
\cmidrule(lr){2-4}
\cmidrule(lr){5-7}
  & \parbox{3em}{$e_{\vec{x}}$ (-)} & \parbox{3em}{$e_{\vec{p}}$ (\si{\metre})} & \parbox{4em}{$e_{\vec{v}}$ (\si{\metre\per\second})}
  & \parbox{3em}{$e_{\vec{x}}$ (-)} & \parbox{3em}{$e_{\vec{p}}$ (\si{\metre})} & \parbox{4em}{$e_{\vec{v}}$ (\si{\metre\per\second})} \\
\midrule
\multicolumn{5}{l}{\textbf{Hovering}
    \sisetup{round-mode=places,round-precision=0,per-mode=power}
    \changed{($\omega \sim \SI{0}{\radian\per\second}$,~$\sigma_\alpha \sim \SI{0}{\radian}$)}}
    \vspace{0.3em}
    \\
    \quad \ac{CV}  & \num{0.559} & \textbf{\num{0.285}} & \num{0.481} & \num{0.559} & \textbf{\num{0.285}} & \num{0.481} \\
  \quad \ac{CA}  & \num{0.540} & \num{0.289} & \num{0.457} & \num{0.539} & \num{0.288} & \num{0.455} \\
    \quad \ac{IMM} & \textbf{\num{0.515}} & \textbf{\num{0.285}} & \textbf{\num{0.429}} & \num{0.517} & \textbf{\num{0.285}} & \num{0.431} \\
\midrule
\multicolumn{5}{l}{\textbf{Flying}
    \sisetup{round-mode=places,round-precision=3,per-mode=power}
    \changed{($\omega \leq \SI[round-precision=0]{1}{\radian\per\second}$,~$\sigma_\alpha \leq \SI{0.005}{\radian}$)}}
    \vspace{0.3em}
    \\
  \quad \ac{CV}  & \num{0.728} & \num{0.271} & \num{0.675} & \num{0.586} & \num{0.240} & \num{0.534} \\
    \quad \ac{CA}  & \num{0.693} & \num{0.275} & \num{0.636} & \num{0.573} & \num{0.250} & \num{0.516} \\
    \quad \ac{IMM} & \num{0.651} & \num{0.270} & \num{0.592} & \textbf{\num{0.531}} & \textbf{\num{0.238}} & \textbf{\num{0.475}} \\ 
\midrule
\multicolumn{5}{l}{\textbf{Maneuvering}
    \sisetup{round-mode=places,round-precision=3,per-mode=power}
    \changed{($\omega \leq \SI[round-precision=1]{2.3}{\radian\per\second}$,~$\sigma_\alpha \leq \SI{0.011}{\radian}$)}}
    \vspace{0.3em}
    \\
  \quad \ac{CV}  & \num{0.919} & \num{0.326} & \num{0.859} & \num{0.670} & \num{0.260} & \num{0.617} \\
  \quad \ac{CA}  & \num{0.864} & \num{0.324} & \num{0.801} & \num{0.647} & \num{0.262} & \num{0.592} \\
    \quad \ac{IMM} & \num{0.816} & \num{0.317} & \num{0.752} & \textbf{\num{0.602}} & \textbf{\num{0.253}} & \textbf{\num{0.542}} \\
\bottomrule
\end{tabularx}
  \sisetup{round-mode=places,round-precision=1,per-mode=power}
  \caption{Comparison of the state estimation methods with and without the measurement uncertainty model for different movement modes of the interceptor.
   \changed{For each metric and movement mode, the best results among the six considered filter variants are highlighted.}
   \changed{The corresponding max. body rotation rate ($\omega$) and orientation uncertainty ($\sigma_\alpha$) for each mode are listed in the brackets. The distance of the target was between \SI{5}{\metre} and \SI{25}{\metre}.}}
  \label{tab:tracking}
\end{table}
}

It may be observed, that the \ac{IMM} outperforms the simpler models in almost all cases, which may be expected as the trajectories generally do not have a constant acceleration or velocity.
The results confirm a better generalization ability of the \ac{IMM}, which is recognized for its efficacy in handling trajectories that undergo mode changes between CV and CA \cite{rong2005imm_survey}.
This is a behavior commonly observed in autonomous drones during scanning or observation tasks.
For all these reasons, we conclude that \ac{IMM} is highly suitable for the task of intercepting non-cooperative \acp{UAV}.

All of the methods exhibited decreased performance with more aggressive interceptor motion, which we attribute to worsened accuracy of the interceptor's self-localization and state estimation, as discussed in sec.~\ref{sec:meas}.
By incorporating the measurement uncertainty model described in that section, the increase in error is significantly mitigated.
\changed{The only exception is for the \textit{hovering} case where the uncertainty is negligible, so the results are similar with and without the model.}

\changed{%
\subsection{Quantitative Evaluation of the Navigation Methods}
\label{sec:res_pn}

We compare the two proposed interception guidance methods with two state-of-the-art methods: the \ac{GPN} control rule from \cite{zhu2017DistributedGuidanceInterception}, and the \ac{MPC} formulation from \cite{srivastava2022intercption}, and two baseline methods: the \ac{PP} and the \ac{LPN}.
To provide a robust comparison, the methods were evaluated using a set of 100 target's trajectories (see Fig.~\ref{fig:trajectories}), with five random starting points of the interceptor per trajectory, and each evaluation lasting 100s, providing almost \SI{14}{\hour} of total flight time.
The trajectories differ in complexity with the fastest 5\% having an average velocity of \SI{5.9}{\metre\per\second} (with \SI{8}{\metre\per\second} maximum) and acceleration of \SI{3.2}{\metre\per\second\squared} (\SI{11}{\metre\per\second\squared} max.).
The overall average velocity and acceleration are \SI{4.1}{\metre\per\second} and \SI{1.4}{\metre\per\second\squared}.
To enable reproducing the results, the dataset is available online\footnote{\href{https://mrs.felk.cvut.cz/towards-interception}{\url{https://mrs.felk.cvut.cz/towards-interception}}}.

The tests were conducted in a custom simulator\footnote{\url{https://github.com/ctu-mrs/mrs_multirotor_simulator}} that supports automated repeated faster-than-real-time simulation with changing starting conditions, which was necessary to make the extensive quantitative evaluation technically feasible.
The simulator is high-fidelity, incorporating aerodynamic drag and full UAV dynamics up to the dynamics of single motors, but does not provide simulation of sensors.
The ground truth state of the target was used as the observation for the navigation algorithms, to allow an examination of each method's limits without favoring a specific detection or tracking approach.


During evaluation, we count the number of \textit{successful interceptions}, which are defined as a passing of the target's center through a circular plane with a \SI{2}{\metre} radius corresponding to the net suspended below the interceptor.
Furthermore, we define the \textit{interception accuracy} as the distance from the net's center during the passing.
Parameter values for \ac{PP}, \ac{LPN}, \ac{GPN}, and \ac{EPN} were selected using grid-search optimization by maximizing the number of trajectories with at least one successful interception as the primary metric, and the number of interceptions per trajectory as the secondary metric (when the primary metric reached the maximum).
We test \ac{GPN} both with the original values from~\cite{zhu2017DistributedGuidanceInterception} (denoted \ac{GPN}\textsubscript{1}) and with the tuned parameters (denoted \ac{GPN}\textsubscript{2}).
Parameters of the \ac{MPC}-based methods cannot be easily optimized in this manner.
For the heading-vector \ac{MPC}~\cite{srivastava2022intercption}, the values from~\cite{srivastava2022intercption} were used except for the dynamic constraints, which were set to match the platform's actual constraints.
For our \ac{MPC} formulation, the parameters were empirically tuned.
The parameter values are listed in Table~\ref{tab:parameters}, and the results are summarized in Table~\ref{tab:simulation_simple}.
}


\begin{table}
\vspace{8pt}
  \centering
  \begin{tabularx}{\linewidth}{ll}
    \toprule
    \textbf{Method} & \textbf{Parameters} \\
    \midrule
    \ac{PP}  & $G = 0.83$ \\
    \ac{LPN} & $G = 19.7$ \\
    \ac{EPN} (ours) & $G = 19.7$, $W = 5.1 \cdot 10^{-2}$ \\
    \ac{GPN}\textsubscript{1} \cite{zhu2017DistributedGuidanceInterception} & $k_1 = 40$,$\phantom{.5}$ $k_2 = 1$,$\phantom{.8}$ $v_r = \SI{-5}{\metre\per\second}$ \\
    \ac{GPN}\textsubscript{2} \cite{zhu2017DistributedGuidanceInterception} & $k_1 = 69.5$, $k_2 = 5.8$, $v_r = \SI{-6.6}{\metre\per\second}$ \\
    \ac{MPC} (ours) & $N = 40$, $\Delta t = \SI{0.2}{\second}$, $\vec{W}_{\vec{e}} = \mathbf{I}_{3 \times 3}$, $\vec{W}_{\vec{u}} = 0.1 \cdot \mathbf{I}_{3 \times 3}$ \vspace{0.3em} \\
    \multirow{2}{*}{\ac{MPC} \cite{srivastava2022intercption}} & $N = 20$, $\Delta t = \SI{0.1}{\second}$, $\vec{W}_{\vec{e}} = \diag(0.5, 0.5, 1, 1)$, \\
    & $\vec{W}_{\vec{u}} = \diag(0.8, 0.8, 0.8, 0.5)$ \vspace{0.3em} \\
    \multirow{2}{*}{all} & $\vec{v}_\text{max} = \bemat{ \SI{8}{\metre\per\second} ~ \SI{8}{\metre\per\second} ~ \SI{4}{\metre\per\second} }\tran$, \\
        & $\vec{a}_\text{max} = \bemat{ \SI{4}{\metre\per\second\squared} ~ \SI{4}{\metre\per\second\squared} ~ \SI{2}{\metre\per\second\squared} }\tran$, $\omega_\text{heading,max} = \SI{2}{\radian\per\second}$ \\
    \bottomrule
  \end{tabularx}
  \caption{
  Parameter values of the navigation algorithms in the experiments.}
  \label{tab:parameters}
\vspace{-0.5em}
\end{table}

\begin{table}
\centering
  \begin{tabularx}{\linewidth}{lrrrrr}
\toprule
    \textbf{Method} & \textbf{Trajs.} & \textbf{N. ints.} & \textbf{$t$ to 1\textsuperscript{st} int.} & \textbf{Int. accur. all / 1\textsuperscript{st}} \\
\midrule
    \ac{PP} & \SI{72}{\percent} & \num{4.40}                                                      & \SI{32.22}{\second} & \SI{1.19}{\metre} / \SI{1.23}{\metre} \\
    \ac{LPN} & \SI{95}{\percent} & \num{12.65}                                                               & \SI{19.73}{\second} & \SI{0.17}{\metre} / \textbf{\SI{0.03}{\metre}} \\
    \changed{\ac{MPC}~\cite{srivastava2022intercption}} & \changed{\SI{95}{\percent}} & \changed{7.77}                                & \changed{\SI{12.83}{\second}} & \changed{\SI{0.38}{\metre} / \SI{0.39}{\metre}} \\
    \ac{GPN}\textsubscript{1} \cite{zhu2017DistributedGuidanceInterception} & \textbf{\SI{100}{\percent}} & \num{10.84} & \SI{21.40}{\second} & \SI{0.17}{\metre} / \SI{0.25}{\metre} \\
    \ac{GPN}\textsubscript{2} \cite{zhu2017DistributedGuidanceInterception} & \textbf{\SI{100}{\percent}} & \num{17.60} & \SI{8.15}{\second} & \SI{0.18}{\metre} / \SI{0.34}{\metre} \\
    \changed{\ac{MPC} (ours)} & \changed{\SI{98}{\percent}} & \changed{16.55}                                & \changed{\SI{11.98}{\second}} & \changed{\SI{0.38}{\metre} / \SI{0.38}{\metre}} \\
    \ac{EPN}\,(ours) & \textbf{\SI{100}{\percent}} & \textbf{\num{24.38}}                                               & \textbf{\SI{5.93}{\second}} & \textbf{\SI{0.16}{\metre}} / \SI{0.04}{\metre} \\
\bottomrule
\end{tabularx}
  \caption{Average results from 500 simulated experiments on 100 different trajectories for each navigation method: the percentage of trajectories with at least one interception attempt, the average number of successful interceptions per trajectory, the time to the first attempt, the interception accuracy per attempt, and the interception accuracy of the first attempt.
  }
\label{tab:simulation_simple}
\end{table}

\changed{%
Based on these experiments, the \ac{EPN} and \ac{GPN}\textsubscript{2} methods have a higher number of total attempts and a significantly lower time to the first attempt than the baseline solutions and the \ac{MPC}-based methods.
}
\ac{EPN} has a similar performance in interception accuracy to \ac{LPN}, whereas \ac{GPN}\textsubscript{2} performs slightly worse.
\changed{The lower performance of the \ac{MPC} methods indicates that distance to the target may not be the best optimization criterion as it leads to less aggressive behavior when the interceptor is lagging closely behind the target.}
The \ac{MPC} planner from~\cite{srivastava2022intercption} is less aggressive and accurate due to the strong limitations enforced by the formulation of the control output.
The main advantage of our approach compared to other methods is the notable improvement for both the reaction times and the number of repeated interceptions.

\changed{%
\subsection{Validation with a Full System}

\begin{figure}
  \centering
  \includegraphics[width=\linewidth]{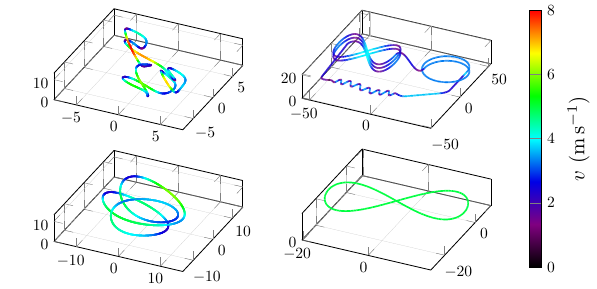}
  \centering 
  \caption{Example trajectories of the target from the evaluation dataset (left column), the trajectory used in the full-system simulation (top right) and the fast trajectory used in the real-world experiments (bottom right).
  Color encodes velocity, dimensions are in meters.
  }
  \label{fig:trajectories}
\end{figure}

To validate performance of the tuned methods when integrated into a full interception system including a real detector, the Gazebo simulator was utilized.}
The system consists of a simulated \ac{LiDAR}, the detection and tracking of flying objects using \ac{VoFOD}~\cite{vrba2023fod}, the \ac{IMM} filter with the \ac{LiDAR} measurement uncertainty model, and an interception navigation algorithm.
\changed{%
  We consider the same navigation methods as in the previous section (with the same parameter values) except the poorly-performing baseline methods \ac{PP} and \ac{LPN}.}
To address the limited field of view of the detector, when the tracking is lost, the planning is interrupted and the interceptor starts rotating in place to re-acquire the detection.
The \acp{UAV} were controlled using the MRS UAV system \cite{mrs_stack}, which allows for control based on desired acceleration or trajectory tracking.


\changed{%
A testing trajectory lasting \SI{330}{\second} was designed that incorporates various modes of target behavior to comprehensively assess capabilities of the system (see Fig.~\ref{fig:trajectories}).
Aggressiveness of the target is lower in comparison to the trajectories from previous section to reduce the number of detection losses (the average velocity and acceleration are \SI{2.7}{\meter\per\second\squared} and \SI{0.7}{\meter\per\second\squared}, the maxima are \SI{4}{\meter\per\second} and \SI{2.5}{\meter\per\second\squared}).}
For each interception algorithm, the target executed the trajectory twice, resulting in approximately \SI{11}{\minute} of data for each method.
We use the evaluation metrics defined in the previous sections except for the time to the first attempt (due to the limited sample size).
In addition, we evaluate the average computation time $\bar{t}_c$, and the detection recall as the percentage of \ac{LiDAR} scans in which the target was detected.
The results are summarized in Table~\ref{table:simulation_gazebo}.

The experiments show that our \ac{EPN} method outperforms the others in the considered metrics with \ac{GPN} and the proposed \ac{MPC} as the two next best methods, which is consistent with the results from sec.~\ref{sec:res_pn}.
\changed{It is worth noting, that the \ac{PN}-based methods have lower and deterministic computational demands, and depend on fewer parameters.
Thus, they are suitable for real-time applications, and they can be easily tuned using optimization, as described in sec.~\ref{sec:res_pn}, which can lead to better results than manual tuning.}
On the other hand, the \ac{MPC} model could be extended to consider obstacles in the environment and improve safety of the interception.
However, optimization-based tuning and obstacle-aware formulation of the \ac{MPC} are beyond the scope of this paper.
%

{
\sisetup{per-mode=repeated-symbol}
\begin{table}
  \centering
  \begin{tabularx}{\linewidth}{Xlllllll}
    \toprule
    \textbf{Method} & \parbox{0.5cm}{\textbf{Num. ints.}} & \parbox{0.8cm}{\textbf{Int. accur.}} & \parbox{0.67cm}{\textbf{Det. recall}} & \parbox{0.5cm}{$e_{\vec{x}}$ (-)} & \parbox{0.5em}{$e_{\vec{p}}$ (\si{\metre})} & \parbox{0.5em}{$e_{\vec{v}}$ (\si{\metre\per\second})} & $\bar{t}_c$\\
    \midrule
    \ac{MPC} \cite{srivastava2022intercption}              & \phantom{0}77 & \SI{1.13}{\metre}           & \SI{51}{\percent} & \num{0.79} & \num{0.38} & \num{0.70} & \SI{4.2}{\milli\second} \\
    \ac{GPN}\textsubscript{1} \cite{zhu2017DistributedGuidanceInterception} & \phantom{0}47           & \SI{1.05}{\metre}           & \SI{47}{\percent} & \num{0.84} & \num{0.35} & \num{0.77} & \SI{130}{\micro\second} \\
    \ac{GPN}\textsubscript{2} \cite{zhu2017DistributedGuidanceInterception} & 189           & \SI{0.79}{\metre}           & \SI{48}{\percent} & \num{0.71} & \num{0.32} & \num{0.64} & \SI{130}{\micro\second} \\
    \ac{MPC}\,(ours)                                        & 188           & \SI{0.84}{\metre}           & \SI{41}{\percent} & \num{0.85} & \num{0.40} & \num{0.75} & \SI{5.1}{\milli\second}  \\
    \ac{EPN}\,(ours)                                        & \textbf{230}  & \textbf{\SI{0.73}{\metre}}  & \SI{47}{\percent} & \num{0.77} & \num{0.35} & \num{0.69} & \SI{130}{\micro\second} \\
    \bottomrule
  \end{tabularx}
  \caption{Results from the \ac{SITL}-simulated interception experiments using the whole control, detection, estimation, tracking, and planning pipeline: the number of successful interceptions, the average interception accuracy, the detection recall, the estimation errors and the computation duration.}
  \label{table:simulation_gazebo}
\end{table}
}



\changed{%
\subsection{Real-world Testing}
Based on the simulated results, we have selected the \ac{IMM} filter with the measurement uncertainty model and the \ac{EPN} planner for the real-world evaluation.
The target was periodically following an eight-shaped trajectory in a $\SI{16}{\metre} \times \SI{40}{\metre}$ area with a constant velocity of $\SI{3}{\metre\per\second}$, an average acceleration of $\SI{0.7}{\metre\per\second\squared}$, and a maximal acceleration of $\SI{1.9}{\metre\per\second\squared}$ (see Fig.~\ref{fig:trajectories}).
The interceptor started in-air at a predefined position near the target's trajectory, and was fully autonomous, running the whole interception pipeline.
To test the limits of the system, the experiment was repeated with an increased velocity of the target to \SI{5}{\metre\per\second} (corresponding to \SI{1.9}{\metre\per\second\squared} average and \SI{4.9}{\metre\per\second\squared} maximal acceleration).
The real-world experiments were only conducted with \ac{EPN} due to technical limitations and safety, so the experiments were reproduced in simulations with the other methods to provide a fair comparison.
The results are presented in Table~\ref{table:rw_experiments} and confirm that our method is capable of intercepting a real-world target flying \SI{5}{\metre\per\second} and significantly outperforms the state of the art.
We have also performed several experiments with a catching net deployed below the interceptor (see Fig.~\ref{fig:intro}), demonstrating the ability of the whole \ac{AAIS} platform to capture the target.
All the experiments are illustrated in the attached video and online\footnote{\href{https://mrs.felk.cvut.cz/towards-interception}{\url{https://mrs.felk.cvut.cz/towards-interception}}}.

}

{
\sisetup{per-mode=repeated-symbol}
\begin{table}
\vspace{8pt}
  \centering
  \begin{tabularx}{\linewidth}{X | rrr | rrr}
    \toprule
                    & \multicolumn{3}{c | }{\textbf{Target's speed: \SI{3}{\metre\per\second}}} & \multicolumn{3}{c}{\textbf{Target's speed: \SI{5}{\metre\per\second}}} \\
    \midrule
    \textbf{Method} & \parbox{0.5cm}{\textbf{Num. ints.}} & \parbox{0.8cm}{\textbf{Int. accur.}} & \parbox{0.8cm}{\textbf{$t$ to 1\textsuperscript{st} int.}} & \parbox{0.5cm}{\textbf{Num. ints.}} & \parbox{0.8cm}{\textbf{Int. accur.}} & \parbox{0.8cm}{\textbf{$t$ to 1\textsuperscript{st} int.}} \\
    \midrule
    \ac{MPC} (ours)*                                        & 7  & \SI{0.82}{\metre}  & \SI{9.82}{\second} & 2  & \SI{0.78}{\metre}  & \SI{11.32}{\second} \\
    \ac{GPN}\textsubscript{2} \cite{zhu2017DistributedGuidanceInterception}* & 11  & \SI{0.68}{\metre}  & \SI{7.82}{\second}    & 3  & \SI{1.39}{\metre}  & \SI{6.34}{\second} \\
    \ac{EPN} (ours)*                                       & 13  & \SI{0.74}{\metre}  & \SI{5.34}{\second} & 4  & \SI{0.99}{\metre}  & \SI{4.52}{\second}  \\
    \ac{EPN} (ours)                                        & 11  & \SI{0.88}{\metre}  & \SI{6.22}{\second} & 3  & \SI{1.03}{\metre}  & \SI{2.54}{\second}  \\
    \bottomrule 
    \multicolumn{7}{l}{\footnotesize *simulated results}
  \end{tabularx}
  \caption{Results from the real-world experiments using \ac{EPN} with the whole pipeline compared to simulated results of the other methods with the same starting conditions and target's trajectory as in the real-world.}
  \label{table:rw_experiments}
\end{table}
}

\vspace{-0.5em}

\section{Conclusions}
Two new planning methods for the safe mid-air interception of intruding \acp{UAV} are proposed in this paper.
The first, dubbed \ac{EPN}, is inspired by \ac{PN} techniques used for missile guidance.
The proposed \ac{EPN} aims to address the requirements of agile \ac{UAV}-based interception.
The second method is formulated as an \ac{MPC} problem optimizing the distance to the target over a prediction horizon.
The methods are compared with two baseline and two state-of-the-art planners, and the \ac{EPN} method shows notable improvements in time to first contact and interception accuracy.
This is complemented with a thorough analysis of state estimation algorithms for target tracking, and a new model for measurement uncertainty that is suitable for sensors onboard a \ac{UAV}, such as the \ac{LiDAR} sensor considered in this work.
As we show in the experiments, incorporating the measurement uncertainty model significantly improves estimation accuracy during agile maneuvers, which is crucial for effective aerial interception.
Finally, the best-performing methods are combined with our previously published \ac{LiDAR}-based onboard detector of flying objects into a first complete \acl{AAIS} that is capable of a real-world interception of agile maneuvering targets.


\vspace{-0.5em}






\bibliographystyle{IEEEtran}
\bibliography{citations}

\begin{thebibliography}{10}
\providecommand{\url}[1]{#1}
\csname url@samestyle\endcsname
\providecommand{\newblock}{\relax}
\providecommand{\bibinfo}[2]{#2}
\providecommand{\BIBentrySTDinterwordspacing}{\spaceskip=0pt\relax}
\providecommand{\BIBentryALTinterwordstretchfactor}{4}
\providecommand{\BIBentryALTinterwordspacing}{\spaceskip=\fontdimen2\font plus
\BIBentryALTinterwordstretchfactor\fontdimen3\font minus \fontdimen4\font\relax}
\providecommand{\BIBforeignlanguage}[2]{{%
\expandafter\ifx\csname l@#1\endcsname\relax
\typeout{** WARNING: IEEEtran.bst: No hyphenation pattern has been}%
\typeout{** loaded for the language `#1'. Using the pattern for}%
\typeout{** the default language instead.}%
\else
\language=\csname l@#1\endcsname
\fi
#2}}
\providecommand{\BIBdecl}{\relax}
\BIBdecl

\bibitem{chamola2021c-uas_survey}
V.~Chamola, P.~Kotesh, A.~Agarwal, {Naren}, N.~Gupta, and M.~Guizani, ``A comprehensive review of unmanned aerial vehicle attacks and neutralization techniques,'' \emph{ADHN}, vol. 111, p. 102324, 2021.

\bibitem{wang2021c-uas_survey}
J.~Wang, Y.~Liu, and H.~Song, ``Counter-unmanned aircraft system(s) ({C-UAS}): State of the art, challenges, and future trends,'' \emph{Aerospace and Electronic Systems Magazine}, vol.~36, no.~3, pp. 4--29, 2021.

\bibitem{ghasri2021accidents}
M.~Ghasri and M.~Maghrebi, ``Factors affecting unmanned aerial vehicles’ safety: A post-occurrence exploratory data analysis of drones’ accidents and incidents in {Australia},'' \emph{Safety Science}, vol. 139, 2021.

\bibitem{vrba2019ral}
M.~{Vrba}, D.~{Heřt}, and M.~{Saska}, ``Onboard marker-less detection and localization of non-cooperating drones for their safe interception by an autonomous aerial system,'' \emph{RA-L}, vol.~4, no.~4, pp. 3402--3409, 2019.

\bibitem{ning2024RealtoSimtoRealApproachVisionBased}
Z.~Ning, Y.~Zhang, X.~Lin, and S.~Zhao, ``A real-to-sim-to-real approach for vision-based autonomous {MAV}-catching-{MAV},'' \emph{Unmanned Systems}, pp. 1--12, Mar. 2024.

\bibitem{vrba2022ras}
M.~Vrba, Y.~Stasinchuk, T.~Báča, V.~Spurný, M.~Petrlík, D.~Heřt, D.~Žaitlík, and M.~Saska, ``{Autonomous capture of agile flying objects using UAVs: The MBZIRC 2020 challenge},'' \emph{RAS}, vol. 149, 2022.

\bibitem{vrba2023fod}
M.~Vrba, V.~Walter, V.~Pritzl, M.~Pliska, T.~Báča, V.~Spurný, D.~Heřt, and M.~Saska, ``On onboard {LiDAR}-based flying object detection,'' 2023, preprint, arXiv 2303.05404, submitted to Transactions on Robotics.

\bibitem{barisic2022fr}
A.~Bari{\v{s}}i{\'{c}}, F.~Petric, and S.~Bogdan, ``Brain over brawn: Using a stereo camera to detect, track, and intercept a faster {UAV} by reconstructing the intruder's trajectory,'' \emph{FR}, vol.~2, no.~1, pp. 222--240, 2022.

\bibitem{vrba2020ral}
M.~{Vrba} and M.~{Saska}, ``Marker-less micro aerial vehicle detection and localization using convolutional neural networks,'' \emph{RA-L}, vol.~5, no.~2, pp. 2459--2466, 2020.

\bibitem{nguyen2019mavnet}
T.~Nguyen, S.~S. Shivakumar, I.~D. Miller, J.~Keller, E.~S. Lee, A.~Zhou, T.~{\"O}zaslan, G.~Loianno, J.~H. Harwood, J.~Wozencraft \emph{et~al.}, ``{MAVNet}: An effective semantic segmentation micro-network for {MAV}-based tasks,'' \emph{RA-L}, vol.~4, no.~4, pp. 3908--3915, 2019.

\bibitem{carrio2018depth}
A.~Carrio, S.~Vemprala, A.~Ripoll, S.~Saripalli, and P.~Campoy, ``Drone detection using depth maps,'' in \emph{IROS}, 2018, pp. 1034--1037.

\bibitem{ning2024bearing}
Z.~Ning, Y.~Zhang, J.~Li, Z.~Chen, and S.~Zhao, ``A bearing-angle approach for unknown target motion analysis based on visual measurements,'' \emph{The International Journal of Robotics Research}, 2024.

\bibitem{srivastava2022intercption}
R.~Srivastava, R.~Lima, and K.~Das, ``Aerial interception of non-cooperative intruder using model predictive control,'' in \emph{ACC}, 2022, pp. 494--499.

\bibitem{wang2020multi_interceptor}
X.~Wang, G.~Tan, Y.~Dai, F.~Lu, and J.~Zhao, ``An optimal guidance strategy for moving-target interception by a multirotor unmanned aerial vehicle swarm,'' \emph{Access}, vol.~8, pp. 121\,650--121\,664, 2020.

\bibitem{moreira2019interception}
M.~Moreira, E.~Papp, and R.~Ventura, ``Interception of non-cooperative {UAVs},'' in \emph{SSRR}, 2019, pp. 120--125.

\bibitem{zhu2017DistributedGuidanceInterception}
B.~Zhu, A.~H.~B. Zaini, and L.~Xie, ``Distributed guidance for interception by using multiple rotary-wing unmanned aerial vehicles,'' \emph{Transactions on Industrial Electronics}, vol.~64, no.~7, pp. 5648--5656, Jul. 2017.

\bibitem{yang2023PolicyLearning}
P.~Yang, S.~Koga, A.~Asgharivaskasi, and N.~Atanasov, ``Policy learning for active target tracking over continuous {$\mathrm{SE}(3)$} trajectories,'' in \emph{Proceedings of The 5th Annual Learning for Dynamics and Control Conference}.\hskip 1em plus 0.5em minus 0.4em\relax PMLR, Jun. 2023, pp. 64--75.

\bibitem{rong2005imm_survey}
X.~Rong~Li and V.~Jilkov, ``Survey of maneuvering target tracking. {Part V.} {Multiple-model methods},'' \emph{AESS}, vol.~41, no.~4, pp. 1255--1321, 2005.

\bibitem{missels}
R.~Yanushevsky, \emph{Modern Missile Guidance}.\hskip 1em plus 0.5em minus 0.4em\relax CRC Press, 2018.

\bibitem{palumbo_missiles}
N.~Palumbo, R.~Blauwkamp, and J.~Lloyd, ``Modern homing missile guidance theory and techniques,'' \emph{Johns Hopkins APL Technical Digest (Applied Physics Laboratory)}, vol.~29, 2010.

\bibitem{verschueren2020acados}
R.~Verschueren, G.~Frison, D.~Kouzoupis, J.~Frey, N.~{van Duijkeren}, A.~Zanelli, B.~Novoselnik, T.~Albin, R.~Quirynen, and M.~Diehl, ``Acados: A modular open-source framework for fast embedded optimal control,'' 2020, preprint, arXiv 1910.13753.

\bibitem{mrs_stack}
T.~Baca, M.~Petrlik, M.~Vrba, V.~Spurny, R.~Penicka, D.~Hert, and M.~Saska, ``The {MRS} {UAV} system: Pushing the frontiers of reproducible research, real-world deployment, and education with autonomous unmanned aerial vehicles,'' \emph{JINT}, vol. 102, no.~26, pp. 1--28, 2021.

\bibitem{HertJINTHW_paper}
D.~{Hert}, T.~{Baca}, P.~{Petracek}, V.~{Kratky}, R.~{Penicka}, V.~{Spurny}, M.~{Petrlik}, M.~{Vrba}, D.~{Zaitlik}, P.~{Stoudek}, V.~{Walter}, P.~{Stepan}, J.~{Horyna}, V.~{Pritzl}, M.~{Sramek}, A.~{Ahmad}, G.~{Silano}, D.~{Bonilla Licea}, P.~{Stibinger}, T.~{Nascimento}, and M.~{Saska}, ``{MRS} drone: A modular platform for real-world deployment of aerial multi-robot systems,'' \emph{JINT}, vol. 108, pp. 1--34, 2023.

\end{thebibliography}

\end{document}